\documentclass[10pt,twocolumn,letterpaper]{article}
\usepackage{simpleConference}
\usepackage{times}
\usepackage{epsfig}

\usepackage{graphicx}
\usepackage{amssymb}
\usepackage{url,hyperref}
\usepackage{float}
\usepackage{amsmath}
\usepackage{caption}
\begin{document}

\title{Deep Sparse Light Field Refocusing}
\author{Shachar Ben Dayan, David Mendlovic and Raja Giryes\\
\\
School of Electrical Engineering, Faculty of Engineering, Tel-Aviv University
\\
bendayan@mail.tau.ac.il, mend@eng.tau.ac.il, raja@tauex.tau.ac.il
\\
}

\maketitle
\thispagestyle{empty}

\begin{abstract}
Light field photography enables to record 4D images, containing angular information alongside spatial information of the scene. One of the important applications of light field imaging is post-capture refocusing. Current methods require for this purpose a dense field of angle views; those can be acquired with a micro-lens system or with a compressive system. Both techniques have major drawbacks to consider, including bulky structures and angular-spatial resolution trade-off. We present a novel implementation of digital refocusing based on sparse angular information using neural networks. This allows recording high spatial resolution in favor of the angular resolution, thus, enabling to design compact and simple devices with improved hardware as well as better performance of compressive systems.  We use a novel convolutional neural network whose relatively small structure enables fast reconstruction with low memory consumption. Moreover, it allows handling without re-training various refocusing ranges and noise levels. Results show major improvement compared to existing methods.
\end{abstract}

\section{Introduction}
\label{sec:intro}

Light field photography has attracted significant attention in recent years due to its unique capability to extract depth without active components  \cite{jeon2018depth,jeon2015accurate,heber2016convolutional}. While 2D cameras only capture the total amount of light at each pixel on the sensor, namely, the projection of the light in the scene, light field cameras also record the direction of each ray intersecting with the sensor in a single capture. Thus, light field images contain spatial and angular information of a given scene. A light field image can be represented as a 4D tensor, carrying the dimensions of the RGB color channels and the multiple angle views alongside the spatial dimensions.

Early methods for light field capturing include multi 2D-cameras array \cite{wilburn2005high}, where each camera captures the scene from a different angle; and sequential imaging \cite{levoy1996light}, where a single camera is shifted and takes multiple images. Both strategies are bulky and not suitable for capturing dynamic scenes. Lytro, a plenoptic camera \cite{ng2005light} captures a light field image in a single exposure. The device is equipped with a micro-lens array located between the sensor and the main lens. This system enables the recording of the direction of the rays arriving at the sensor plane. However, since the sensor is meant to capture both spatial and angular information, there is an inherent tradeoff between the captured spatial and angular resolution.

Other methods such as coded masks \cite{veeraraghavan2007dappled} or coded aperture \cite{liang2008programmable} based light field also suffer from the above trade-off and produce images with low spatial resolution. Recent approaches suggest using compressive systems for light field acquisition \cite{marwah2013compressive,chen2017light,gupta2017compressive,nabati2018fast}, where the camera records a coded projection of the light rays intersecting with the sensor. The light field images are then recovered from their projections using methods such as sparse coding and deep learning. However, the more angular information there is to compress, the more artifacts the output images have, resulting in poor resolution and signal-to-noise ratio (SNR).

\begin{figure}
\centering
\includegraphics[width=\linewidth]{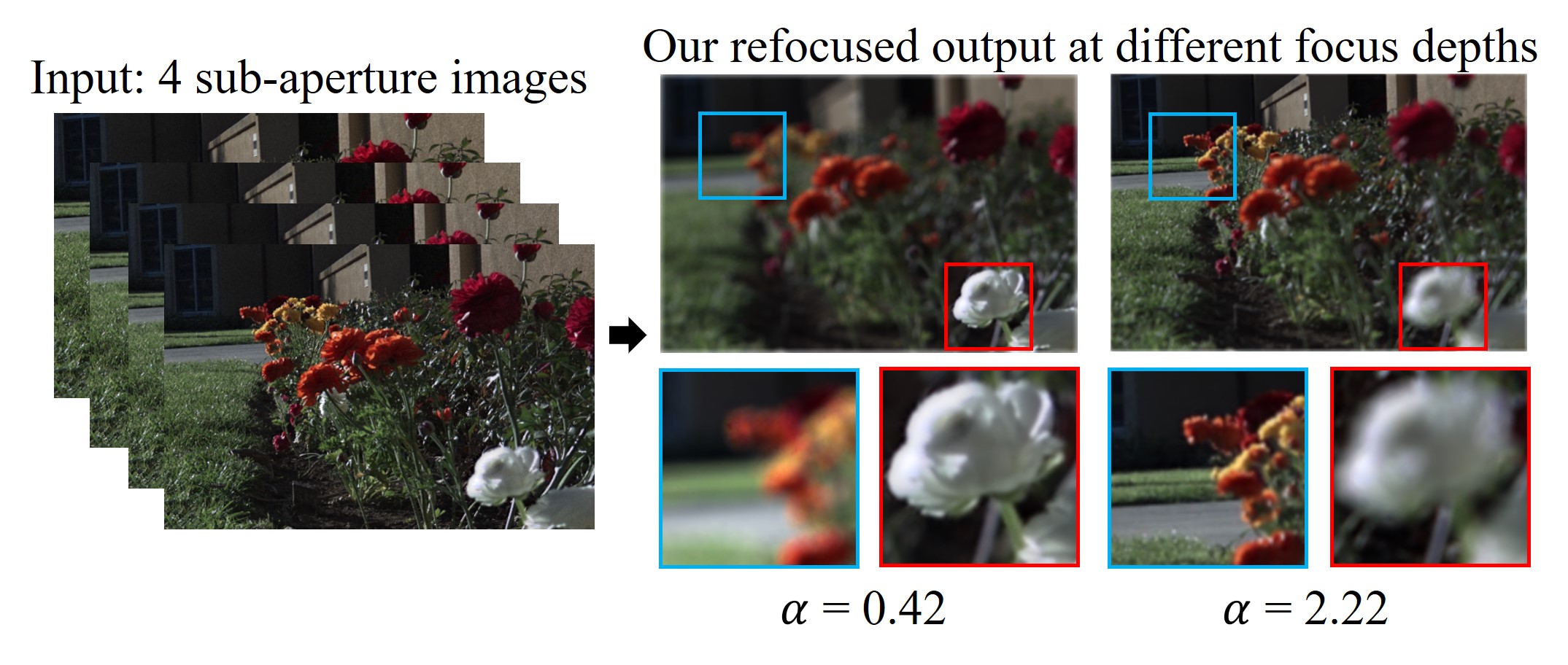}
\caption{Our proposed method of light field refocusing needs only 4 sub-aperture input views to output a good refocused image at a specific depth.}
\label{fig:teaser}
\end{figure}

{\bf Contribution.} Given the above, it is important to perform light field applications such as refocusing, which is the focus of this work, relying only on sparse angular data. We propose a technique for post-capture refocusing based on sparse light field data: only 2x2 sub-aperture views. To this end, we exploit the power of convolutional neural networks (CNNs). The four input views are fed into a CNN, which outputs a refocused image according to a given focus plane (see Fig.~\ref{fig:teaser}). To the best of our knowledge, we are the first to suggest an end-to-end light field refocusing based on deep learning. 

Our approach shows improved performance in terms of PSNR, SSIM and computational time, compared to the current existing methods, which are of two kinds: direct refocusing, and indirect synthesis methods that synthesize from a small number of views a full light field to generate from it the refocused image.
Our trained network, RefocusNet, exhibits very good generalization performance across different sensors. We demonstrate that after training it on a dataset from one sensor, it can be used with another different light field system without the need of a new training.

\section{Related Work}
\label{chap:2}

While there are many works on refocusing \cite{wang2018deeplens, matzen2019synthetic, suwajanakorn2015depth, barron2015fast, park2017unified}, this work focuses on light field based refocusing. The basic light field refocusing algorithm relies on shifting the sub-aperture views according to the desired focus plane and the disparity of each sub-aperture view and then averaging the shifted images \cite{levoy2004synthetic,vaish2004using}. This method has been demonstrated on light field images captured by a Lytro camera \cite{ng2005light} and was recently generalized to tilt-shift photography \cite{alain2019interactive}. However, this method requires a dense angular sampled light field, or otherwise the resulted images will contain artifacts (see Section \ref{chap:algorithm}). 
Several methods locate and fix the aliasing caused by the lack of angular information \cite{xiao2014aliasing, lee2016depth}.

Another body of works uses the Fourier Slice Theorem \cite{bracewell1956strip,marichal2011fast,fu2015implementing}. The Fourier Slice Digital Refocusing method \cite{ng2005fourier} proves that spatial light field refocusing is equal to computing the Fourier transform of a 4D light field, selecting a relevant 2D slice and then performing 2D inverse Fourier transform. Yet, it does not perform well on sparse angular data. 
A recent strategy \cite{le2019fourier} represents light fields efficiently using Fourier Disparity Layers (FDL). This compact representation can be used for sub-aperture views rendering and direct refocusing.

Other methods suggest increasing the angular resolution of a light field by reconstructing additional sub-aperture views and produce the refocused images by shifting and averaging the sub-aperture images \cite{mitra2012light}. With the grown popularity of deep learning methods in computer vision tasks, a CNN has been trained to reconstruct a 7x7 light field from 3x3 given angles, while using EPIs (epipolar plane images) \cite{wu2017light}. While this method outperforms state-of-the-art methods, they indicate that the algorithm will not perform well on inputs with less than 3x3 angle views. The work of Kalantari \textit{et al.} \cite{kalantari2016learning} uses two sequential CNNs to synthesize an 8X8 light field from 4 (2x2) corner sub-aperture images. Although this method performs well on sparse angular data, the fact that each view needs to be created separately is time-consuming, which makes the algorithm not suitable for real-time applications. This is a major drawback when relying on light field synthesis methods  \cite{wang2018end,penner2017soft,srinivasan2017learning, ivan2019synthesizing} for refocusing;
instead, direct refocusing can be performed using the given sub-aperture views, similar to image fusion networks \cite{sun2019rtfnet, krispel2020fuseseg}.


\section{Mathematical Background}
\label{chap:3}
\noindent {\bf Light Field Photography.}
Levoy \textit{et al.} \cite{levoy1996light} introduced the modeling of a 4D light field as a two-plane parameterization, which simplifies the problem formulation. We briefly describe it here. 
The common light field device is usually equipped with a micro-lens array located between the sensor and the main lens, which is placed at the aperture plane.
Given the sensor plane $(x,y)$, the aperture plane $(u,v)$, and the distance between them $F$ (focal length), we may say that the intensity of each point $(x,y)$ on the sensor plane equals to the integration of all the light rays moving through the whole aperture reaching this point:

\begin{equation}
\hspace{-0.09in} {E}_{F} (x,y) =\frac{1}{F^2}\iint {L}_{F} (x,y,u,v) A(u,v)cos^4 {\phi}\mathrm{d}u\mathrm{d}v. \end{equation}

${E}_{F}(x,y)$ is the intensity at point $(x,y)$ on the sensor plane (known as image plane), ${L}_{F}$ is the radiance along the ray from $(u,v)$ to $(x,y)$, $A$ is the aperture function (1 inside the aperture and 0 outside), and $\phi$ is the angle between the image plane normal and the light ray. Let
\begin{equation}
\overline{{L}_{F}} (x,y,u,v) = {L}_{F} (x,y,u,v) cos^4 {\phi},
\end{equation}
then if we assume that ${E}_{F}$ is zero outside the aperture boundaries, we get the simplified form:
\begin{equation}
\label{eq:E_F_equality}
{E}_{F} (x,y) =\frac{1}{F^2}\iint \overline{{L}_{F}}  (x,y,u,v) \mathrm{d}u\mathrm{d}v. \end{equation}

\noindent {\bf Digital Refocusing.}
An important light field application is post-capture refocusing. As 2D images only contain spatial information, we have to choose the focus plane before capturing the image and adjust the aperture accordingly. In light field photography, the angular information is also available, which allows us to change the focus plane also after capturing. 

\begin{figure}
\centering
\includegraphics[width=\linewidth]{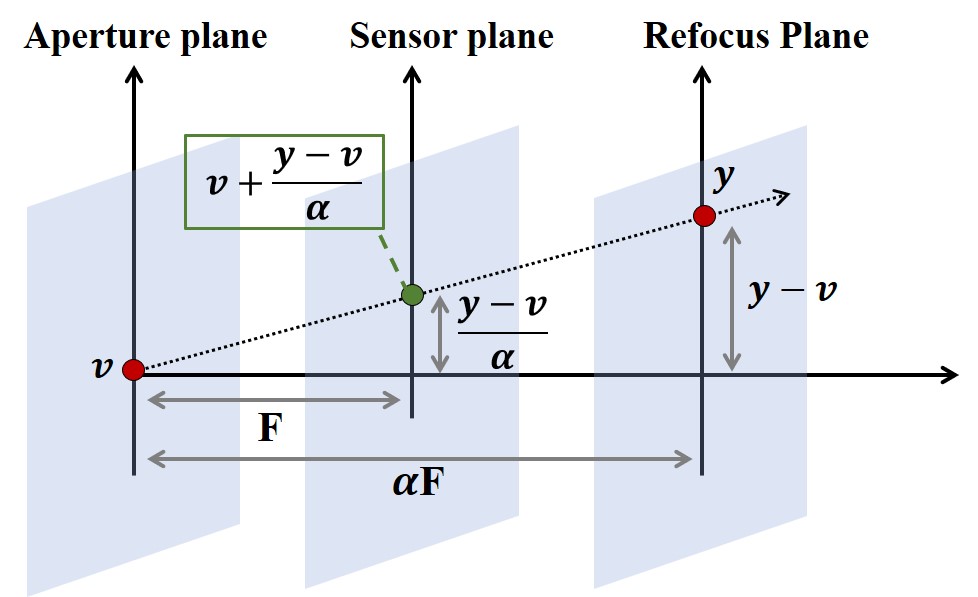}
\caption{Changing the sensor plane artificially to get a refocused image. The new plane is located at a distance of $F'$=$\alpha$$F$ from the aperture plane.}
\label{fig:refocus}
\end{figure}

To produce a refocused light field image, one has to artificially change the current sensor plane. As all possible focus planes are parallel, we need to multiply the current plane, located at distance $F$ from the aperture, by a factor $\alpha$ to get the new focus plane, $F'$. 
Fig.~\ref{fig:refocus} shows that 
using triangles similarity, we can compute the new intersection point of the ray with the sensor plane. 
Substituting $(x,y)$ with the new coordinates and $F' = \alpha F$ in \eqref{eq:E_F_equality}, we get:

\begin{eqnarray}
\label{eq:1}
    && \hspace{-0.2in} {E}_{(\alpha F)} (x,y) =\\ \nonumber
    && =\frac{1}{\alpha^2 F^2}\iint \overline{{L}_{F}}  (u+\frac{x-u}{\alpha},v+\frac{y-v}{\alpha},u,v) \mathrm{d}u\mathrm{d}v.
\end{eqnarray}

 From \eqref{eq:1} we may conclude that practically, creating a refocused image is done by shifting the sub-aperture images, each by a different factor determined by the focus factor and view location on the rectangular grid, followed by averaging the shifted images (each corresponding to a different sub-aperture view). 
This can be explained from an optical point of view: each sub-aperture image represents a different pinhole on the aperture plane. Since light field is the summation of rays intersecting at a specific pixel on the sensor, we sum the sub-aperture images. In order to embody the location of each pinhole on the aperture plane in the refocused image, we shift the sub-aperture images.

The refocused image model in Eq.~\eqref{eq:1} is continuous. However, a single light field image is a discrete set of 2D images. In order to imitate this continuity and create a smooth image, the light field set needs to be as dense as possible, i.e., be with a high angular resolution. On the other hand, because of the spatial-angular resolution trade-off mentioned above, we would like to be able to decrease the angular resolution as much as possible and still get a high quality refocused image. Our proposed method, described next, targets this challenge.

\section{Proposed Method}
\label{chap:algorithm}
Our goal is to address the spatial-angular resolution trade-off and create a light field refocusing program that operates in real-time with low computational complexity. Thus, we propose a convolutional neural network (CNN) based approach for refocusing from sparse light field. 

\begin{figure}
\centering
\includegraphics[width=\linewidth]{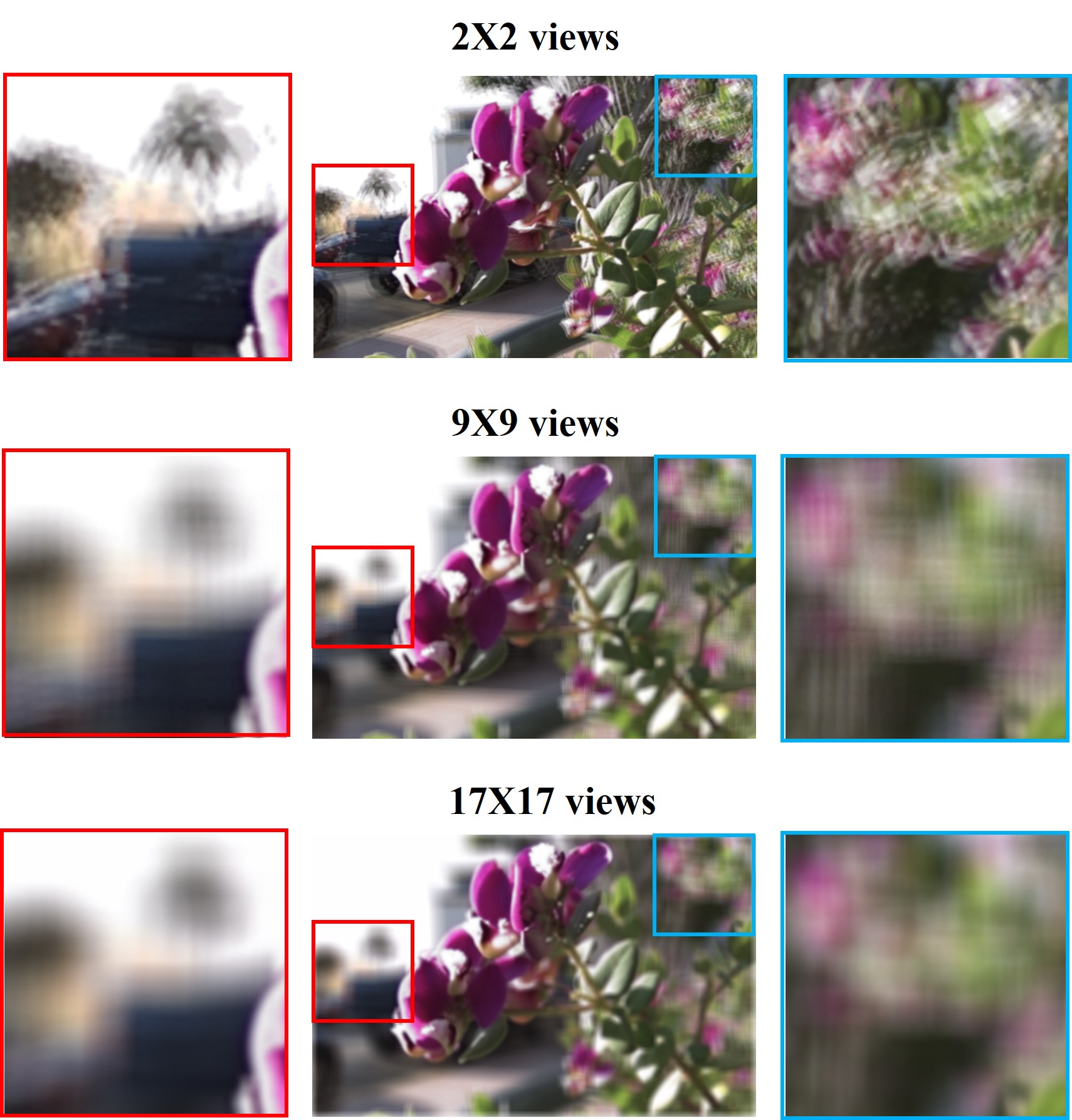}
\caption{Demonstration of conventional light field refocusing based on a varying number of sub-aperture views. The ghosting artifacts are clearly seen in the 2x2 grid result. Also, minor artifacts are still seen in the 9x9 grid result (note the tree and the edge of the car). However, we can barely see any artifacts in the 17x17 grid.}
\label{fig:3}
\end{figure}

\noindent {\bf Sparse Angular Data Refocusing.}
Based on the assumption that we only have 4 input views arranged on a 2x2 grid, we first check what happens when we refocus an image only from these views, without any priors. We shift each view according to its location on the grid and the target focus factor, and then sum the shifted views and divide them by 4. We use the code in \cite{WinNT2}. As expected, the resulted image has noticeable ghosting artifacts around objects that are not on the selected focus plane (see Fig.\ref{fig:3}). 

Note that the differences around the region in focus (the central flower) in Fig.~\ref{fig:3}, between the refocusing based on 2x2 views and the refocusing based on a higher number of views are negligible; the reason is that when we refer to an object at the focus plane, the shift of the sub-aperture images equals to zero, and all the shifted sub-aperture images are the same and averaging them results in the same image, making the object appear sharp and clear. On the contrary, when looking at objects outside of the focus plane, we expect them to appear blurred. Theoretically, if we would have used an infinite number of sub-aperture views, we would get an average of infinitesimal shifts, which looks like a blur to the human eye; yet, sparse light field grid is not enough to produce a smooth, continuous refocused image and therefore the objects outside of the focus plane contain artifacts.

\begin{figure*}
\centering
\includegraphics[width=\linewidth]{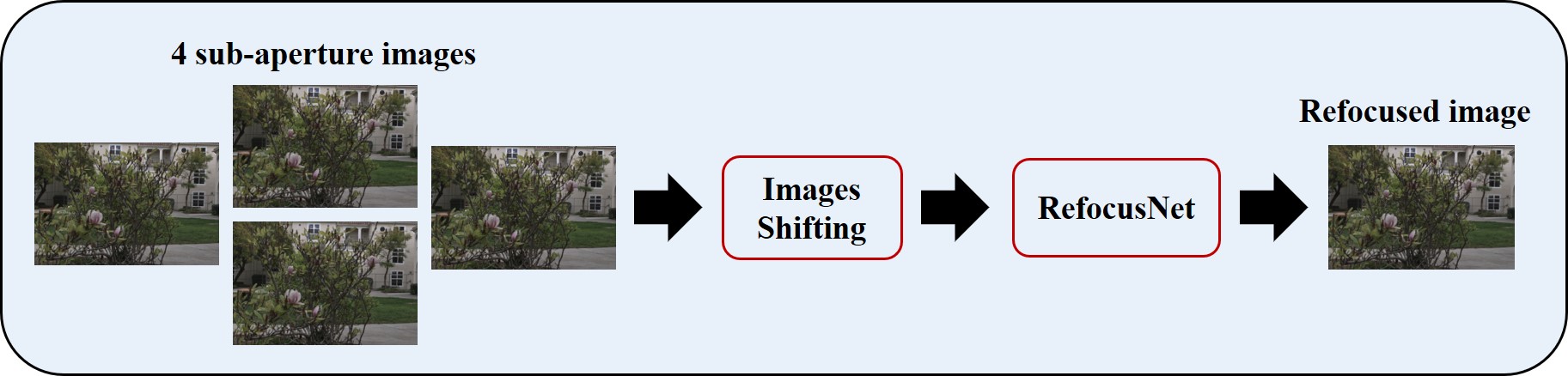}
\caption{Flow chart of the proposed algorithm. Given 4 sub-aperture images, we shift the views according to their locations on the rectangular light field grid and the randomly chosen focus factor. Then we feed them into a CNN that outputs a refocused RGB image.}
\label{fig:4a}
\end{figure*}

Therefore, in order to produce a continuous refocused image when given a sparse light field data, we had to find a way to add more angular information to the process without actually creating these angles. For this purpose, we employ convolutional neural networks, which have both a strong expressive power on images and generalization abilities. The input to our CNN is a tensor of 2X2 angle views, shifted in advance according to the views' location on the grid and the randomly chosen focus factor.  The ground truth is a refocused RGB image generated from a 17X17 light field grid using shifting and averaging. The output of the network is a refocused RGB image (see Fig.~\ref{fig:4a}).

\noindent {\bf Ground Truth and Input Generation.}
The ground truth images that are used for our network training are created by shifting and averaging the sub-aperture views. We use the Lytro Illum camera datasets in \cite{WinNT, kalantari2016learning}, given in the form of lenslet images that can be separated to 14x14 2D angle views. However, we only use 9x9 out of 14x14 given views, to avoid the corruption appearing near the borders of the sub-aperture views located at the edges of the grid, due to vignetting effect. 
In general, vignetting effects can be removed in pre-processing (as was done with our prototype camera images) and are not the focus of this work.

Still, we notice minor artifacts and bold edges in the resulting image, because the light field is not dense enough. Therefore, we decide to up-sample the light field grid by adding more samples in-between the given views, while maintaining the original disparity range of the 9x9 light field. We generate the in-between views using the algorithm in \cite{kalantari2016learning}. We train their network to get an input of 2X2 and output a 3X3 grid. Then we run the algorithm on 2X2 segments of the whole 9X9 field, creating a 17x17 field (Fig. 1 in appendix A).

In addition, in order to make the artificial refocused ground truth appear as similar as possible to an optically generated refocused image, we only use part of the 17X17 input views, which are arranged in a circular shape. Thus, appearing as close as can be to the shape of the camera aperture. This assumption turns out to be correct because the views on the edges of the rectangular grid are often corrupted with black pixels that appear in the process of creating them from the raw light field data.
 In total, we used 241 views out of the 289 views (see Fig. 2 in appendix A). 
 
 The four ($2\times2$) sub-aperture images, used as the network's input, are in the form of a rhombus. We choose this shape, and not a rectangle \cite{kalantari2016learning}, as it matches a prototype of a light field camera developed in our lab, described next.  The chosen views are close enough to the center of the grid but still far enough to take advantage of the disparity between them (see Fig. 2 in appendix A).

\textbf{Demo light field camera.} As mentioned earlier, in Lytro camera \cite{ng2005light} a single sensor captures both the spatial and angular information. As a result of the inherent tradeoff between them, the sub-aperture views have a low spatial resolution - 376 x 541. In order to improve the spatial resolution, a light field camera prototype was built in our lab using off-the-shelf components. This prototype favors the spatial resolution (1024 x 1280) over the angular resolution (2 x 2), and a research effort is invested in solving the challenges that arise from the low angular resolution. The camera pin-holes are designed in the shape of a rhombus and a rotating cover exposes a single pin-hole each time. The final setup, which captures 4 sub-aperture images at once, is based on deep reconstruction of compressed light fields \cite{nabati2018fast}. To show the generalization ability of our algorithm across optical devices, we tested it both on images captured by a Lytro camera as well as images captured by our device. 

\begin{figure}
\centering
\includegraphics[width=\linewidth]{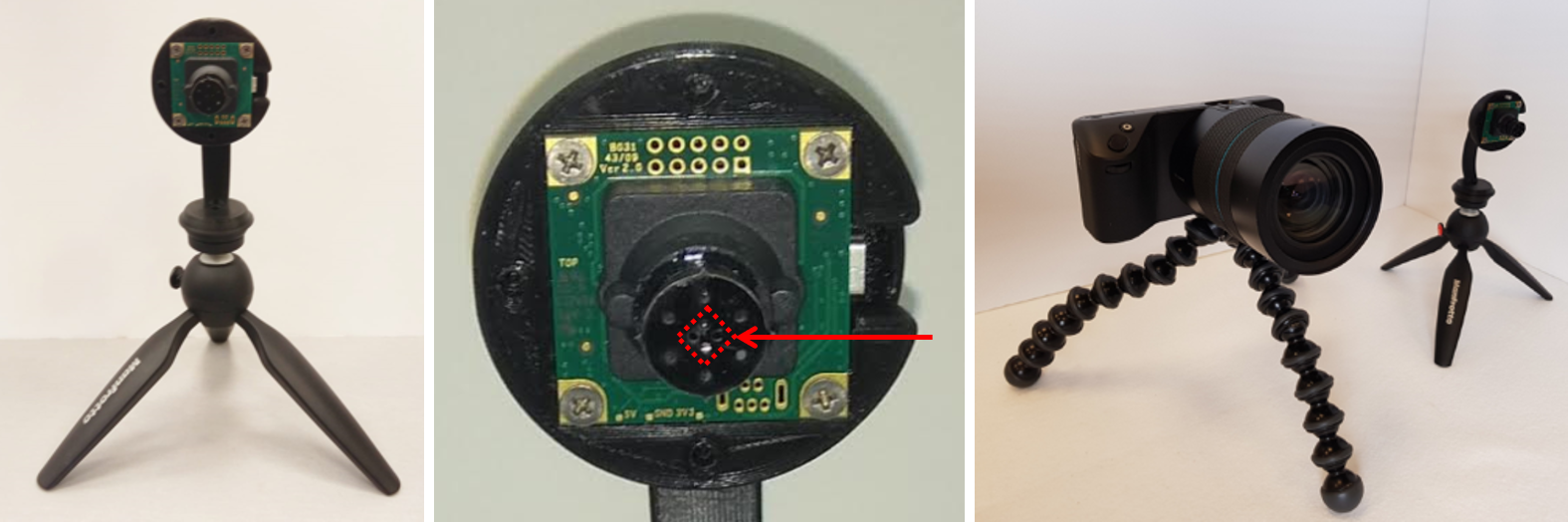}
\caption{A light field camera prototype built in our lab. Left: the entire setup. Middle: close-up, where the 4 pin-holes are marked in red. Right: our prototype has compact dimensions in comparison to Lytro camera.}
\label{fig:6.5}
\end{figure}

\begin{figure*}
\centering
\includegraphics[width=0.8\linewidth]{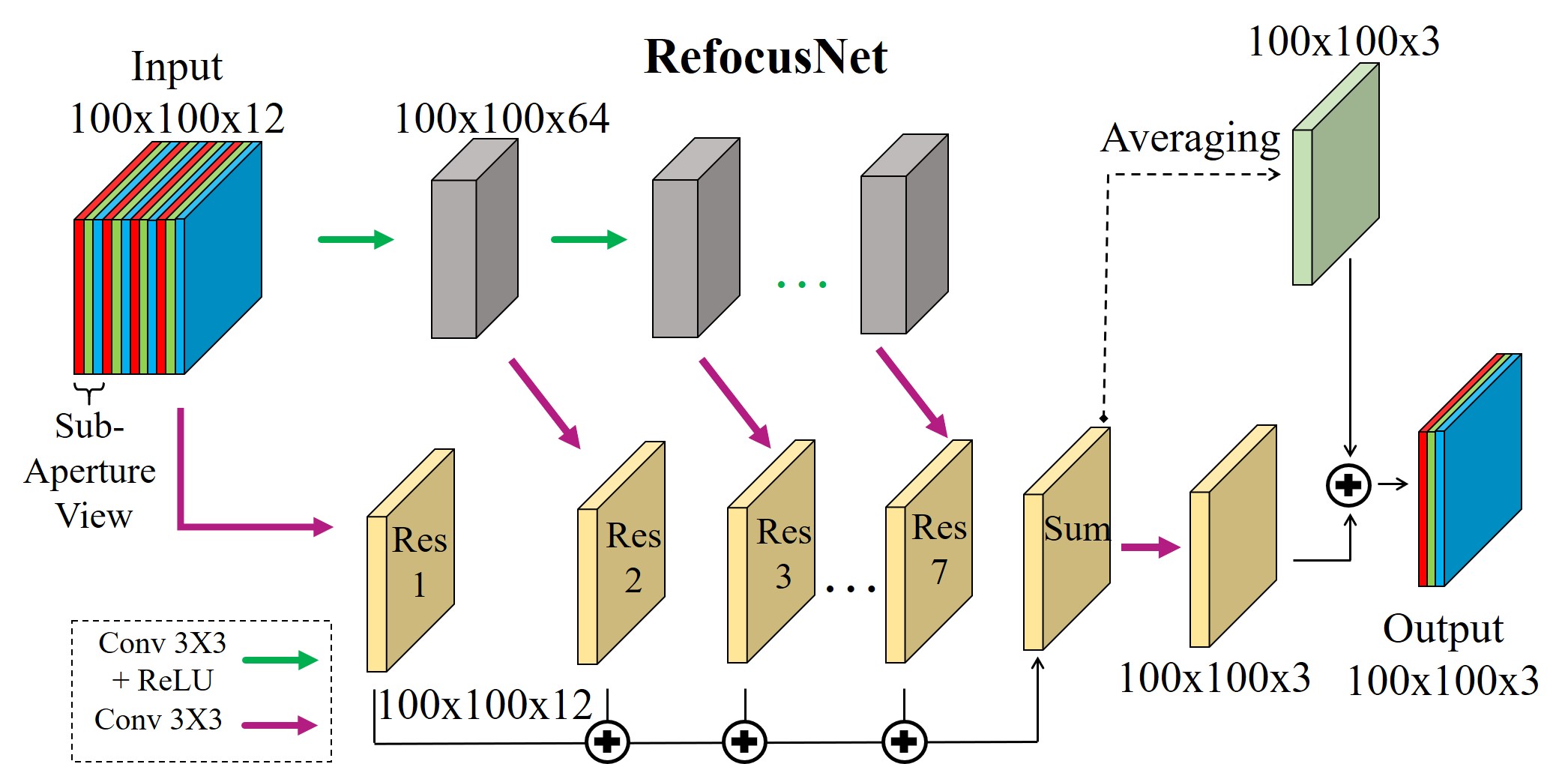}
\caption{Illustration of the \textbf{RefocusNet} architecture. The input consists of 4 sub-aperture views concatenated as a tensor. The network has 2 parallel trajectories; the output of each layer in the first trajectory is used as an input to a matching layer in the second trajectory.}
\label{fig:7}
\end{figure*}

\noindent {\bf Network Architecture.}
Our proposed refocusing network, \textit{RefocusNet}, is presented in Fig~\ref{fig:7}. 
It is a fully convolutional architecture with residual connections, and has two parallel trajectories. Each layer's output from the first trajectory is the input to the matching layer in the second trajectory. The summation of the seven residuals goes under two operations: averaging of the 12 channels; and a 3x3 convolution. These two outputs are then summed to get the network output. We train it using the $\ell_{1}$ loss as it is more robust to outliers in the training data compared to $\ell_2$. 

The contribution of this specific architecture to our problem is that it preserves the parts in the refocused image that we wish to preserve and repairs the parts that need smoothing e.g. the parts with the ghosting artifacts. While this network is based on an existing architecture named DenoiseNet \cite{remez2017deep, remez2018deep}, we have made several important changes in it to improve the performance on both clean and noisy data as shown in appendix B in several ablation studies that we have conducted.

\noindent {\bf Computational efficiency.} In 
appendix B we show that an important advantage of our method is that it has low memory usage, and thus can be used in embedded systems.

\section{Experiments}
\label{chap:4}

We turn to compare our technique to other methods:  FDL representation based refocusing \cite{le2019fourier}, and two light field synthesis methods \cite{kalantari2016learning,wu2017light}, which are used to generate a full field of 17x17, followed by the basic shifting and averaging refocusing algorithm \cite{levoy2004synthetic,vaish2004using,ng2005light}.
We show our results on the Lytro dataset \cite{WinNT} combined with the dataset of \cite{kalantari2016learning}, and on our demo camera images. 
Several ablation studies conducted on our method appear in appendix B.

\subsection{Training setup}
We train the network on images captured with a Lytro camera, using the Stanford Lytro dataset \cite{WinNT} and the dataset provided in \cite{kalantari2016learning}. We divide the data into a training set of 300 images, a validation set of 20 images and a test set of 20 images. The original size of the light field images from both datasets is 376x541x3. However, as described above, we use a synthesizing algorithm \cite{kalantari2016learning} to generate a denser light field of 17x17 angle views, which crops the pixels on the sides of the images to prevent synthesizing artifacts. Therefore we get smaller input images of size 332x497x3. 

We train the network on randomly chosen patches of size 100x100, for  $\sim$300 epochs with a batch-size of 32 and a total of 9,600 patches per each epoch that are randomly selected from the images. 
The color channels of each of the input angles are concatenated, i.e., the input tensor in the 4-input-views case is of size [32, 100, 100, 12]. 
We use the Adam optimizer \cite{kingma2014adam} and the weights are initialized with Xavier initialization \cite{glorot2010understanding}.
We set the learning rate to be with an exponential decay in a staircase form, where the initial learning rate is set to 0.0005. During inference, we exploit the fact that our network is fully convolutional and therefore can be applied to the whole image at once, although it was trained on patches.

Another issue that emerged during test time is border effects. When using the traditional method \cite{levoy2004synthetic,vaish2004using,ng2005light} of shifting and averaging the sub-aperture views, a linear interpolation is involved in the process to fill in the missing pixels values. However, since at the edges of the image there are no pixel neighbors and the information is missing, the interpolation performed causes border effects. Empirically we have found that there are border effects within 6 pixels from the edges and therefore we remove these parts. 

\noindent {\bf Focus parameter setting.} For the refocusing done based on the traditional method \cite{levoy2004synthetic,vaish2004using,ng2005light}, we used the code in \cite{WinNT2}, that uses the parameter \textit{pixels} to set the focus point. The relation between the parameter $\alpha$ from Eq. \eqref{eq:1} and \textit{pixels} is:
\begin{equation}
\label{eq:6}
\alpha = \frac{1}{1-\textit{pixels}}
\end{equation}
The values of the parameter \textit{pixels} were chosen empirically in the range [-1.50,1.30] with spaces of 0.05. These values were selected to match the existing range in the Lytro dataset. Clearly, this range is not suitable for images taken with our demo camera; more details on the latter are provided hereafter.

\noindent {\bf Refocusing Noisy Data.}
We extend the method described above to noisy light field data. The goal is to generate denoised refocused images when we are given only noisy data. 
To this end, we have added Gaussian noise with a standard deviation distributed uniformly at random from the range [0, 0.08] to the training data, while the ground truth image is left noiseless. The test is performed on data with the maximal added noise - Gaussian noise with a standard deviation of 0.08. 

\noindent {\bf Refocusing light field images captured with a demo camera.}
As mentioned earlier, to show that our algorithm is independent of the input data type, we tested it also on light field images captured with a prototype designed at our lab. We transferred the network trained for the Lytro dataset to the dataset acquired in our lab, to show the generalization ability of the network. However, for making the transfer possible, the images should have the same scale and therefore we rescaled the demo images by a factor of 0.4 before feeding them to the network; also, the focus range was changed to a new range. The focus parameter \textit{pixels} was determined empirically within the range [-6,6] with spaces of 0.05.

\subsection{Results}
\label{sec:Results}

\begin{figure*}[h]
\centering
\includegraphics[width=0.9\linewidth]{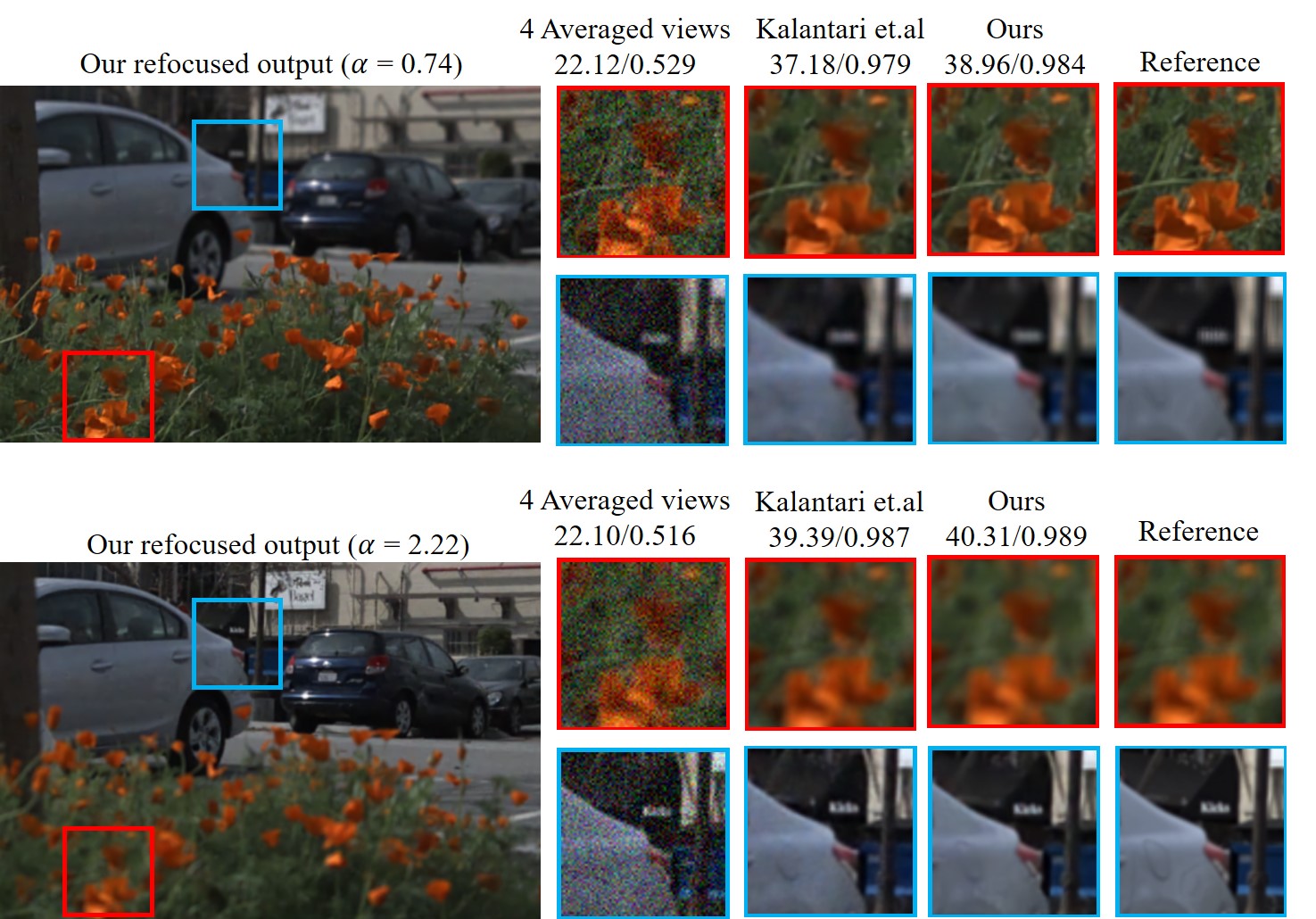}
\caption{Comparison of refocusing and denoising results  (PSNR/SSIM) between the traditional method of shifting and averaging with 4 given sub-aperture views  (\cite{ng2005light,levoy2004synthetic,vaish2004using}), Kalantari \textit{et al.} \cite{kalantari2016learning} and our RefocusNet. Our algorithm cleans the noise in addition to refocusing on the different focus planes, while the traditional method fails in denoising, and the results of Kalantari \textit{et al.} contain some noise (see the car in the blue patch of focus 2.22). }
\label{fig:10}
\end{figure*}



\begin{table}
\centering
\begin{tabular}{|c c c c|}
\hline
\multicolumn{4}{|c|}{\textit{}{Input views: 2x2}} \\ 
\hline
                                                             & \textbf{PSNR} & \textbf{SSIM} & \textbf{Time (Sec)} \\ \hline
RefocusNet                                                    & \textbf{46.15}         & \textbf{0.997}         & 0.023               \\ \hline
FDL \cite{le2019fourier}                                                         & 34.10          & 0.938         & $\sim$4               \\ \hline

\begin{tabular}[c]{@{}c@{}}Kalantari \textit{et al.} \cite{kalantari2016learning}\end{tabular} & 43.50         & 0.990         & 3,198                \\ \hline
\begin{tabular}[c]{@{}c@{}}Wu \textit{et al.} \cite{wu2017light}\end{tabular} & 34.49          & 0.939         & 1,556                \\ \hline
\end{tabular}

\caption{Results of light field refocusing with 2x2 given input views using suggested RefocusNet, compared to FDL refocusing \cite{le2019fourier} and two synthesis methods: Kalantari \textit{et al.} \cite{kalantari2016learning} and Wu \textit{et al.} \cite{wu2017light}. Since we run FDL refocusing \cite{le2019fourier} on a CPU, we used the computational times reported in their paper for time estimation on a GPU.}
\label{table:1}
\end{table}

\begin{table}
\centering
\begin{tabular}{|c c c c|}
\hline
\multicolumn{4}{|c|}{\textit{}{Input views: 3x3}} \\ 
\hline
                                                             & \textbf{PSNR} & \textbf{SSIM} & \textbf{Time (Sec)} \\ \hline
RefocusNet                                                    & \textbf{48.60}         & \textbf{0.998}         & 0.028               \\ \hline
FDL \cite{le2019fourier}                                                         & 34.57          & 0.939         & $\sim$4               \\ \hline

\begin{tabular}[c]{@{}c@{}}Kalantari \textit{et al.} \cite{kalantari2016learning}\end{tabular} & 44.46          & 0.993         & 2,670                \\ \hline
\begin{tabular}[c]{@{}c@{}}Wu \textit{et al.} \cite{wu2017light}\end{tabular} & 36.35          & 0.946          & 1,536                \\ \hline
\end{tabular}

\caption{Results of light field refocusing with 3x3 given input views using suggested RefocusNet, compared to FDL refocusing \cite{le2019fourier} and two synthesis methods: Kalantari \textit{et al.} \cite{kalantari2016learning} and Wu \textit{et al.} \cite{wu2017light}.}
\label{table:2}
\end{table}

\begin{table}
\centering
\begin{tabular}{|c c c c|}
\hline
\multicolumn{4}{|c|}{\textit{}{Noisy Data}} \\ 
\hline
                                                             & \textbf{PSNR} & \textbf{SSIM} & \textbf{Time (Sec)} \\ \hline
RefocusNet                                                    & \textbf{40.95}         & \textbf{0.990}         & 0.022               \\ \hline
FDL \cite{le2019fourier}                                                         & 29.66         & 0.777         & $\sim$4               \\ \hline

\begin{tabular}[c]{@{}c@{}}Kalantari \textit{et al.} \cite{kalantari2016learning}\end{tabular} & 39.45         & 0.974         & 3,763                \\ \hline
\end{tabular}
\caption{Results of light field refocusing on noisy data with 2x2 input views using suggested RefocusNet, compared to FDL refocusing \cite{le2019fourier} and a synthesis method: Kalantari \textit{et al.} \cite{kalantari2016learning}.}
\label{table:3}
\end{table}

{\bf Lytro data.} Table \ref{table:1} presents the numerical evaluation of our averaged results on the test set. Our suggested RefocusNet has the best results both in PSNR and SSIM compared to FDL representation based refocusing \cite{le2019fourier} and the synthesis methods \cite{kalantari2016learning} and \cite{wu2017light}. The computation time is about 2 orders of magnitude faster than FDL refocusing \cite{le2019fourier} and 5 orders of magnitude faster than the synthesis methods. Since Wu \textit{et al.} \cite{wu2017light} claim that their synthesis algorithm does not perform well when given an angular resolution lower than 3x3, we also trained RefocusNet on an input of 3x3 views and compared the results to the other methods (table \ref{table:2}). The results show that the performance of all methods improves when increasing the angular resolution, as expected, and our method still has the best performance.

Next, we test our RefocusNet on the noisy test set (described above). 
Table \ref{table:3} summarizes the results and demonstrates that RefocusNet has the best results, both in PSNR and SSIM also in this case. Since the traditional refocusing method is based on data averaging, which also has a denoising effect \cite{singh2014performance}, we had to check whether just averaging the 4 given input views achieves satisfying results in the denoising and refocusing tasks. Fig.~\ref{fig:10} shows the performance of RefocusNet on a noisy test image, compared to the result of 4 views averaging, Kalantari \textit{et al.} \cite{kalantari2016learning} and the clean ground truth. Observe that the shifting and averaging alone leaves noticeable noise, while Kalantari \textit{et al.} removes most of the noise but not all of it, and its computational time is not applicable for real-time applications. However, our algorithm performs better denoising while also refocusing the image, in real-time.

\begin{figure}[h]
\centering
\includegraphics[width=\linewidth]{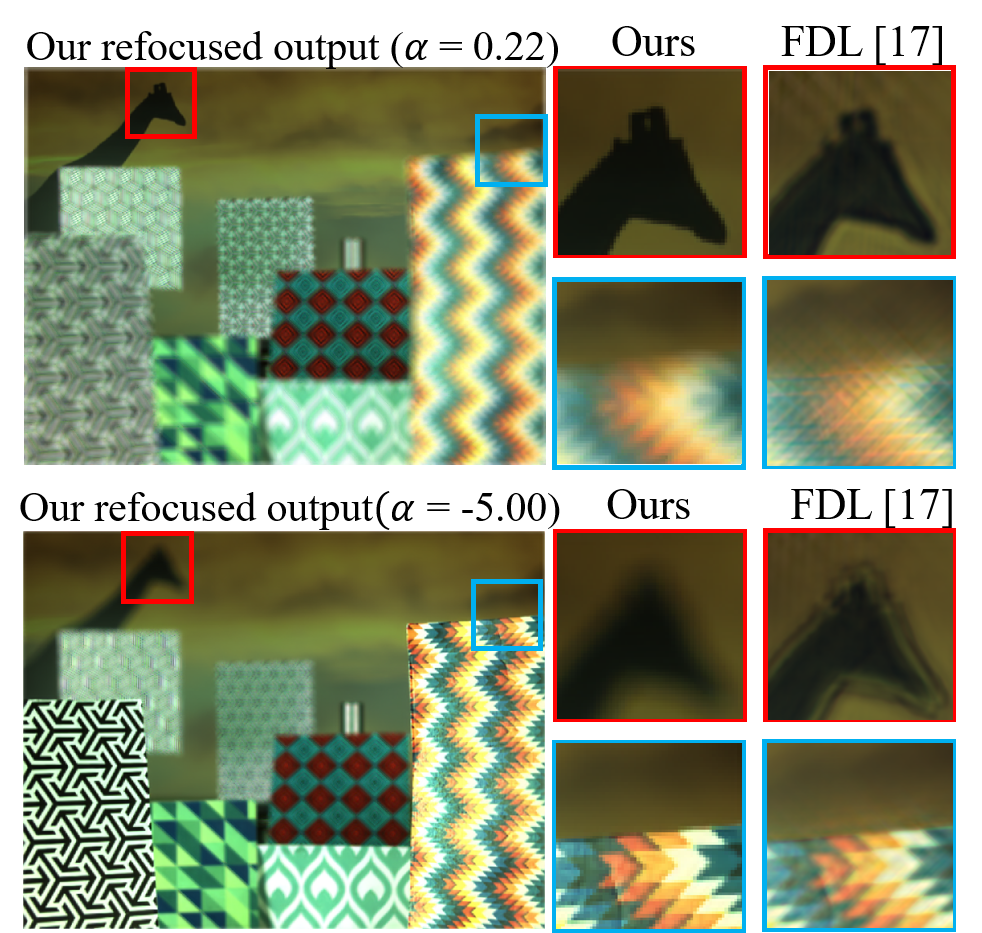}
\caption{Refocusing results on an image taken with a camera demo built in our lab, comparing our method with FDL refocusing \cite{le2019fourier}. Note the noticeable artifacts in the background and near the giraffe's head in their result, while our method successfully smooths these areas. }
\label{fig:11}
\end{figure}

{\bf Real system setup.} Finally, we test our refocusing network on light field images that were taken with a camera demo built in our lab. Notice that no re-training is required and we use the pre-trained network of the Lytro dataset. This demonstrates the generalization ability of our proposed approach. Fig.~\ref{fig:11} exhibits our refocusing results compared to FDL refocusing \cite{le2019fourier}, which contain noticeable artifacts. The successful refocusing demonstrated in this example (and others in appendix C) shows the generalization ability of our technique to different capturing devices. 


\section{Conclusion}
\label{chap:5}
In this work, we have proposed an algorithm for digital light field refocusing based on sparse angular information with convolutional networks. To the best of our knowledge, this is the first suggested method for end-to-end light field refocusing based on deep learning.
The CNN gets four sub-aperture views and creates a refocused image according to a given focus. We have proposed a novel refocusing network that exhibits very good results in real-time. Experiments show that our method significantly improves the quality of refocusing compared to the previous approaches, with a fast implementation and low memory. Moreover, our approach generalizes well across different sensors without the need to fine-tune it. 

\bibliographystyle{abbrv}
\bibliography{LF_Refocusing}
\end{document}


\renewcommand\thesection{\Alph{section}}
\title{Appendices}
\author{}
\maketitle
\thispagestyle{empty}


\section{Additions to section 4: Proposed Method}
\begin{figure}[H]
\centering
\includegraphics[width=0.5\linewidth]{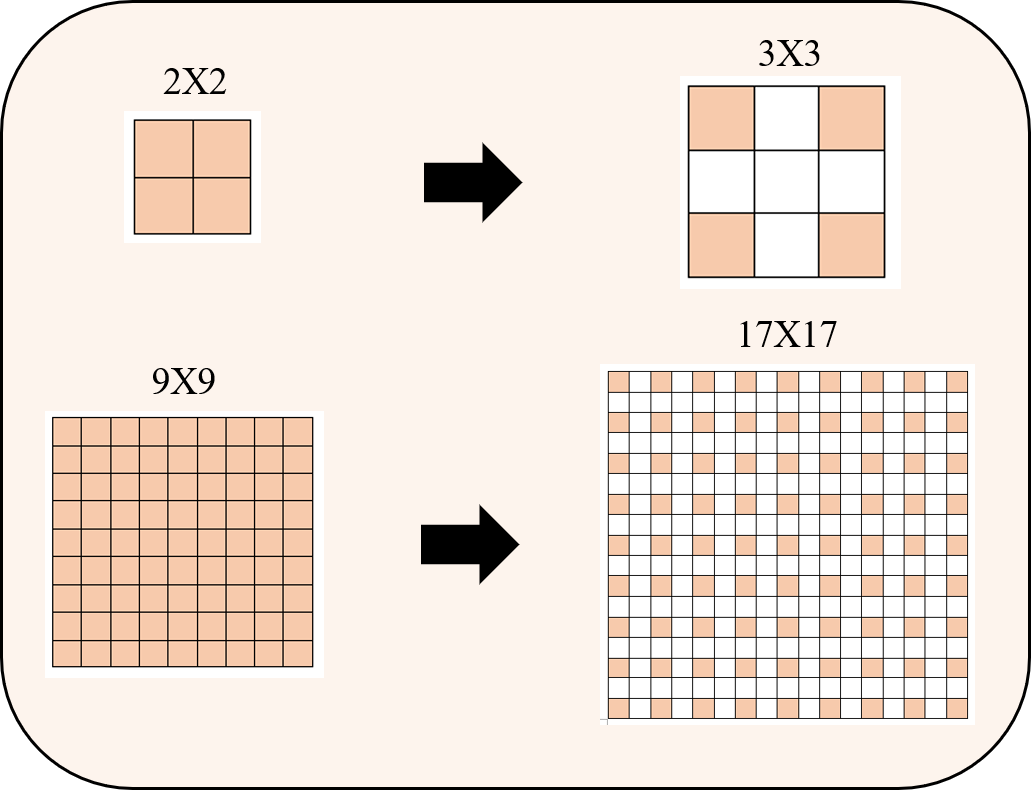}
\caption{Illustration of creating a denser light field grid by computing in-between views of the given images. We divide the 9x9 grid to blocks of 2x2 and feed them to a synthesizing net \cite{kalantari2016learning}, which computes a 3x3 block from each of them. Then we arrange the blocks together to create a 17x17 rectangular grid.}
\label{fig:1}
\end{figure}

As explained in this section, we up-sample the light field grid by adding more samples in-between the given views, while maintaining the original disparity range of the 9x9 light field.
This was done using the algorithm in \cite{kalantari2016learning}. We train their network to get an input of 2X2 and output a 3X3 grid. Then we run the algorithm on 2X2 segments of the whole 9X9 field, creating a 17x17 field. This can be seen in figure \ref{fig:1}. We refer to the 17x17 synthetic light field data as our initial dataset. 

\begin{figure}[H]
\centering
\includegraphics[width=0.5 \linewidth]{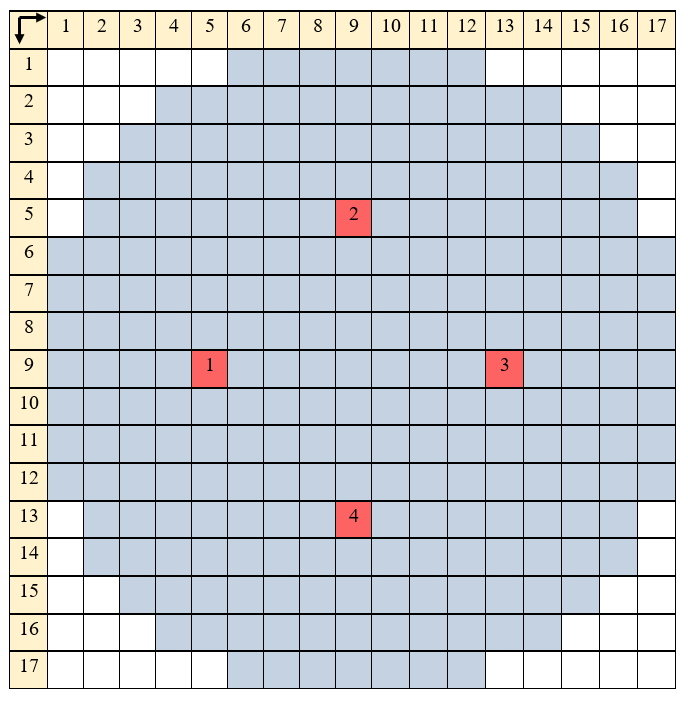}
\caption{Illustration of the 241 views that are used to compute the ground truth image, arranged in the form of a circle to appear close to a circular aperture in the optical device. The 4 views that are used as an input to the network are marked in red and numbered according to their order in the input tensor.}
\label{fig:2}
\end{figure}

In addition, we only use part of the 17X17 input views, which are arranged in a circular shape. Thus, appearing as close as can be to the shape of the camera aperture. In total, we used 241 views out of the 289 views (see figure \ref{fig:2}). Regarding the network's input, we choose four views which are arranged in the shape of a rhombus. As described in the main paper, this was done to match the pin-holes of a light field camera demo developed in our lab.

\section{Additions to section 5: Experiments}
\subsection{Network Architectures}
As we are not aware of any previous work that suggests a CNN that performs light field based refocusing, we compare our model to three popular CNN architectures. The networks we evaluate are described below, and their performance on the test set is summarized in tables \ref{table:1} and \ref{table:2}.

\textbf{1. Simple CNN:} 
A simple CNN with 7 layers, with constant filter sizes of [3,3] and 128 channels. Each layer is followed by ReLU and batch-normalization \cite{ioffe2015batch}.

\textbf{2. ResNet:} 
A modified implementation of ResNet \cite{he2016deep} with four residual blocks, each block consists of two layers. In addition, there is a convolutional layer before the residual blocks and a convolutional layer after the residual blocks. We use ReLU and batch-normalization within each layer, except in the last layer. All the layers have constant filter sizes of [3,3] and 64 channels.

\textbf{3. DenseNet:}
A modified implementation of DenseNet \cite{huang2017densely}, with two residual blocks, each block consists of two sub-blocks. Each sub-clock consists of two convolutional layers, one with a filter size of [1,1] and the other one with a filter size of [3,3]. The first sub-block has 6 channels in each layer, and since the outputs are concatenated the output has 12 channels. The second sub-block has 12 channels in each layer, resulting in an output of 24 channels. After each dense block, there is a transition layer, reducing the number of channels from 24 to 6. Each layer is followed by ReLU and batch-normalization. 

The loss function used to train all architectures is:
 \begin{equation}
 {\ell}_{refocus} =\overset{B}{\underset{j=1}{\sum}}\| \hat{{P}_{j}}-{P}_{j} {\|}_{1},
 \end{equation}
where B is the batch size, $\hat{{P}_{j}}$ is the refocused reconstruction and ${P}_{j}$ is the ground truth patch. We choose the $\ell_{1}$ loss as it is more robust to outliers in the training data compared to $\ell_2$.

\begin{table}[H]
\centering
\begin{tabular}{|c c c c|}
\hline
\multicolumn{4}{|c|}{Clean Data - Input views: 2x2}                                                                       \\ \hline

                                                             & \textbf{PSNR} & \textbf{SSIM} & \textbf{Time (Sec)} \\ \hline
RefocusNet                                                    & \textbf{46.15}         & \textbf{0.997}         & 0.023               \\ \hline
ResNet\cite{he2016deep}                                                       & 41.87         & 0.994         & 0.031               \\ \hline
\begin{tabular}[c]{@{}c@{}}Simple CNN\end{tabular}       & 40.57         & 0.994         & 0.043               \\ \hline
DenseNet \cite{huang2017densely}                                                     & 37.09         & 0.982         & 0.013               \\ \hline
\end{tabular}

\caption{Results of light field refocusing with 2x2 given input views using four different CNN architectures.}
\label{table:1}
\end{table}

\begin{table}[H]
\centering
\begin{tabular}{|c c c c|}
\hline
\multicolumn{4}{|c|}{\textit{}{Noisy Data - Input views: 2x2}} \\ 
\hline
                                                             & \textbf{PSNR} & \textbf{SSIM} & \textbf{Time (Sec)} \\ \hline
RefocusNet                                                    & \textbf{40.95}         & \textbf{0.990}         & 0.022               \\ \hline
ResNet\cite{he2016deep}                                                        & 38.87         & 0.987         & 0.031               \\ \hline
\begin{tabular}[c]{@{}c@{}}Simple CNN\end{tabular}       & 38.90         & 0.988         & 0.044               \\ \hline
DenseNet \cite{huang2017densely}                                                       & 33.88         & 0.970         & 0.013               \\ \hline

\end{tabular}
\caption{Results of light field refocusing on noisy data with 2x2 given input views using four different CNN architectures.} 
\label{table:2}
\end{table}

\subsection{Computational efficiency}

An important advantage of our method is that it has low memory usage, and therefore can be used for applications on mobile devices, where the storage is very limited.  
Table \ref{table:3} summarizes the storage properties of each of our architectures and Kalantari \textit{et al.} \cite{kalantari2016learning}, when the later is divided into two ways of computation during test time: parallel computation and sequential computation. The parameter \textbf{network size} includes the sizes of the weights and biases of the networks. The parameter \textbf{memory} includes the size of occupied memory required for the activations in inference time. For Kalantari \textit{et al.} , during \textbf{parallel computation} the memory is multiplied by $17^2$ while during \textbf{sequential computation} the computational time is multiplied by $17^2$, where $17^2$ is the number of light field images synthesized in the process.

As can be seen in Table \ref{table:3},  our most efficient network, RefocusNet, requires about 760 times less memory than Kalantari \textit{et al.} in the parallel case and about 2.6 times less memory in the sequential case; however, in the latter case their algorithm requires 7 orders of magnitude of computational time more than ours.

\begin{table}[H]
  \centering
\begin{tabular}{|cccc|}
\hline
\textbf{Architecture}        & \textbf{\begin{tabular}[c]{@{}c@{}}Net. Size\\  (MB)\end{tabular}} & \textbf{\begin{tabular}[c]{@{}c@{}}Memory\\  (MB)\end{tabular}} & \textbf{\begin{tabular}[c]{@{}c@{}} Time at Test \\ (Secs.)\end{tabular}} \\ \hline
RefocusNet                   & 0.94                                                              & 50.16                                                           & 0.023                                                                                  \\ \hline
Simple CNN                   & 3.02                                                              & 84.48                                                           & 0.043                                                                                  \\ \hline
ResNet \cite{he2016deep}                       & 1.22                                                              & 84.48                                                           & 0.031                                                                                  \\ \hline
DenseNet\cite{huang2017densely}                    & 0.02                                                              & 15.84                                                           & 0.013                                                                                  \\ \hline
Kalantari \textit{et al.} \cite{kalantari2016learning}-  Parallel  & 6.58                                                              & 38,148                                                        & 3,198                                                                                   \\ \hline
Kalantari \textit{et al.} \cite{kalantari2016learning}- Sequential& 6.58                                                              & 132                                                          & 924,222                                                                                 \\ \hline
\end{tabular}
\caption{Comparison of different network architectures and Kalantari \textit{et al.} \cite{kalantari2016learning} in terms of memory used (including activations) and computational time for an image of size 332x497x3.}
\label{table:3}
\end{table}

\subsection{Refocusing Ablation Study}
\label{subsection:Ablation Study}
\subsubsection{\textbf{Number of Inputs}} 
We turn to explore the influence of the number of sub-aperture views given as an input to the neural network on the refocusing quality. We train it on (i) 2 input views, (ii) 4 input views shaped like a rhombus, (iii) 4 input views shaped like a rectangle and (iv) 8 input views shaped as the union of the two combinations of the 4 views. We also compute the refocused image using the traditional shifting and averaging method \cite{levoy2004synthetic,vaish2004using,ng2005light} as a reference. 

\begin{table}[H]
\centering
\begin{tabular}{|ccc|}
\hline
\multicolumn{3}{|c|}{\textit{2 Horizontal Views}}                                                          \\ \hline
\textbf{Architectures} & \multicolumn{1}{l}{\textbf{PSNR/SSIM}} & \multicolumn{1}{l|}{\textbf{Time (Sec)}} \\ \hline
RefocusNet             & \multicolumn{1}{l}{42.17 / 0.994}      & 0.021                                    \\ \hline
Traditional \cite{levoy2004synthetic,vaish2004using,ng2005light}              & \multicolumn{1}{l}{31.63 / 0.884}      & 14.58                                     \\ \hline
\multicolumn{3}{|c|}{\textit{4 Rectangular Views}}                                                             \\ \hline
\textbf{Architectures} & \textbf{PSNR/SSIM}                     & \textbf{Time (Sec)}                      \\ \hline
RefocusNet             & 44.74 / 0.996                          & 0.023                                    \\ \hline
Traditional \cite{levoy2004synthetic,vaish2004using,ng2005light}              & 37.82 / 0.948                          & 15.94                                     \\ \hline
\multicolumn{3}{|c|}{\textit{4 Rhomboid Views}}                                                         \\ \hline
\textbf{Architectures} & \textbf{PSNR/SSIM}                     & \textbf{Time (Sec)}                      \\ \hline
RefocusNet             & 46.15 / 0.997                          & 0.023                                    \\ \hline
Traditional \cite{levoy2004synthetic,vaish2004using,ng2005light}              & 35.34 / 0.941                          & 16.11                                     \\ \hline
\multicolumn{3}{|c|}{\textit{8 Views}}                                                                     \\ \hline
\textbf{Architectures} & \textbf{PSNR/SSIM}                     & \textbf{Time (Sec)}                      \\ \hline
RefocusNet            & 48.21 / 0.998                          & 0.026                                    \\ \hline
Traditional \cite{levoy2004synthetic,vaish2004using,ng2005light}              & 39.40 / 0.971                          & 23.70                                     \\ \hline
\end{tabular}
\caption{Sub-aperture views ablation study. Comparing between different input sizes: 2 sub-aperture views, 4 sub-aperture views shaped as a rectangle, 4 sub-aperture views shaped as a rhombus, and 8 sub-aperture views. The results are averaged over 20 test images.}
\label{table:4}
\end{table}

As can be seen in table \ref{table:4}, the highest improvements in the PSNR values of RefocusNet compared to the traditional approach occur in the use-cases of 2 horizontal views and 4 rhomboid views (around 10 dB). Our network's performance increases constantly when increasing the number of sub-aperture input views, and when changing the shape of 4 input views from rectangle to rhombus. The computational time at test time increases with the number of sub-aperture views but in a negligible range of a few hundredths of a second. We can also see that, as expected, when applying the traditional refocusing method of shifting and averaging, increasing the number of sub-aperture views results in a consistent increase in the PSNR values as well as the computational time (due to the added information). 
We may conclude that for a refocusing neural network, the optimal number of sub-aperture views needed as input is 4 rhomboid views; since in these ranges of PSNR (45 dB and higher), the improvements are barely seen to the human eye. Using 8 input views will require higher angular resolution of the capturing device and higher computational complexity. Yet, all of this may lead to a marginal benefit.

\subsubsection{\textbf{Number of Layers}}
We also tested the influence of the network's depth on its performance. Table \ref{table:5} shows that increasing the number of layers from 7 to 9 and 11 only improves the performance by marginal values and increases the computational time. Considering the trade-off between performance and computational time and the principal of vanishing returns, we decided to choose an architecture with 7 layers. 

\begin{table}[H]
\centering
\begin{tabular}{|ccc|}
\hline

\textbf{\begin{tabular}[c]{@{}c@{}}Number of\\   layers\end{tabular}} & \textbf{PSNR/SSIM}   & \textbf{Time {[}sec{]}} \\ \hline
3                                                                     & 43.44/0.995          & 0.012                   \\ \hline
5                                                                     & 45.39/0.996          & 0.017                   \\ \hline
7                                                                     & 46.15/0.997          & 0.023                   \\ \hline
9                                                                     & 46.53/0.997          & 0.029                  \\ \hline
11                                                                     & 46.23/0.997          & 0.035                  \\ \hline
\end{tabular}
\caption{Performance of RefocusNet as a function of the network's depth.}
\label{table:5}
\end{table}

\subsubsection{\textbf{Contribution of Skip Connections}}
\begin{figure}[h]
\centering
  \begin{subfigure}[b]{0.4\columnwidth}
    \includegraphics[width=\linewidth]{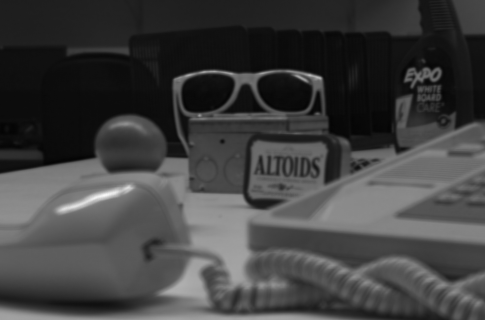}
    \caption{\nth{1} residual layer}
  \end{subfigure}
    \begin{subfigure}[b]{0.4\columnwidth}
    \includegraphics[width=\linewidth]{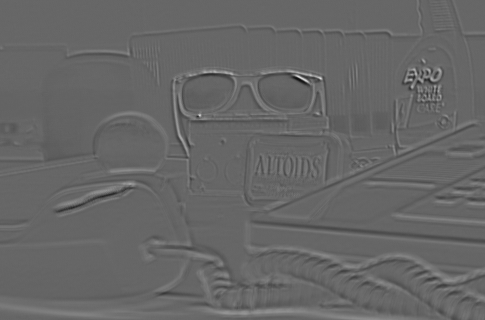}
    \caption{\nth{3} residual layer}
  \end{subfigure}
  \begin{subfigure}[b]{0.4\columnwidth}
    \includegraphics[width=\linewidth]{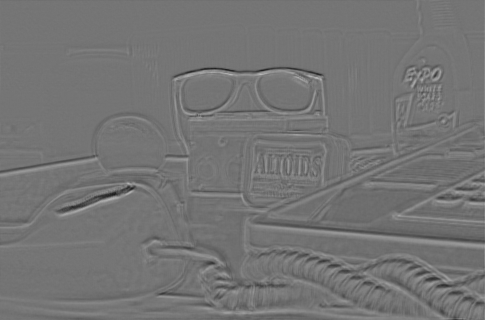}
    \caption{\nth{5} residual layer}
  \end{subfigure}
    \begin{subfigure}[b]{0.4\columnwidth}
    \includegraphics[width=\linewidth]{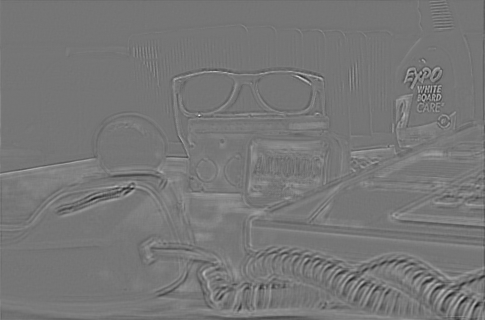}
    \caption{\nth{7} residual layer}
  \end{subfigure}  
\caption{Visualization of the 1st, 3rd, 5th and 7th residual layers.}
\label{fig:3}
\end{figure}

In order to estimate the contribution of the skip connections in RefocusNet, we trained an almost identical network - but without the skip connections. The performance during inference decreased significantly - more than 10 dB (from 46.21 to 35.33), which shows us that adding the residual layers is crucial to the network performance. The computational time, however, also decreased to 0.017 seconds.  A visualization of the residual layers can be seen in figure \ref{fig:3}.




\subsubsection{\textbf{RefocusNet Variations}}
\label{subsection:Variations}
\begin{figure}[H]
\centering
\includegraphics[width=\linewidth]{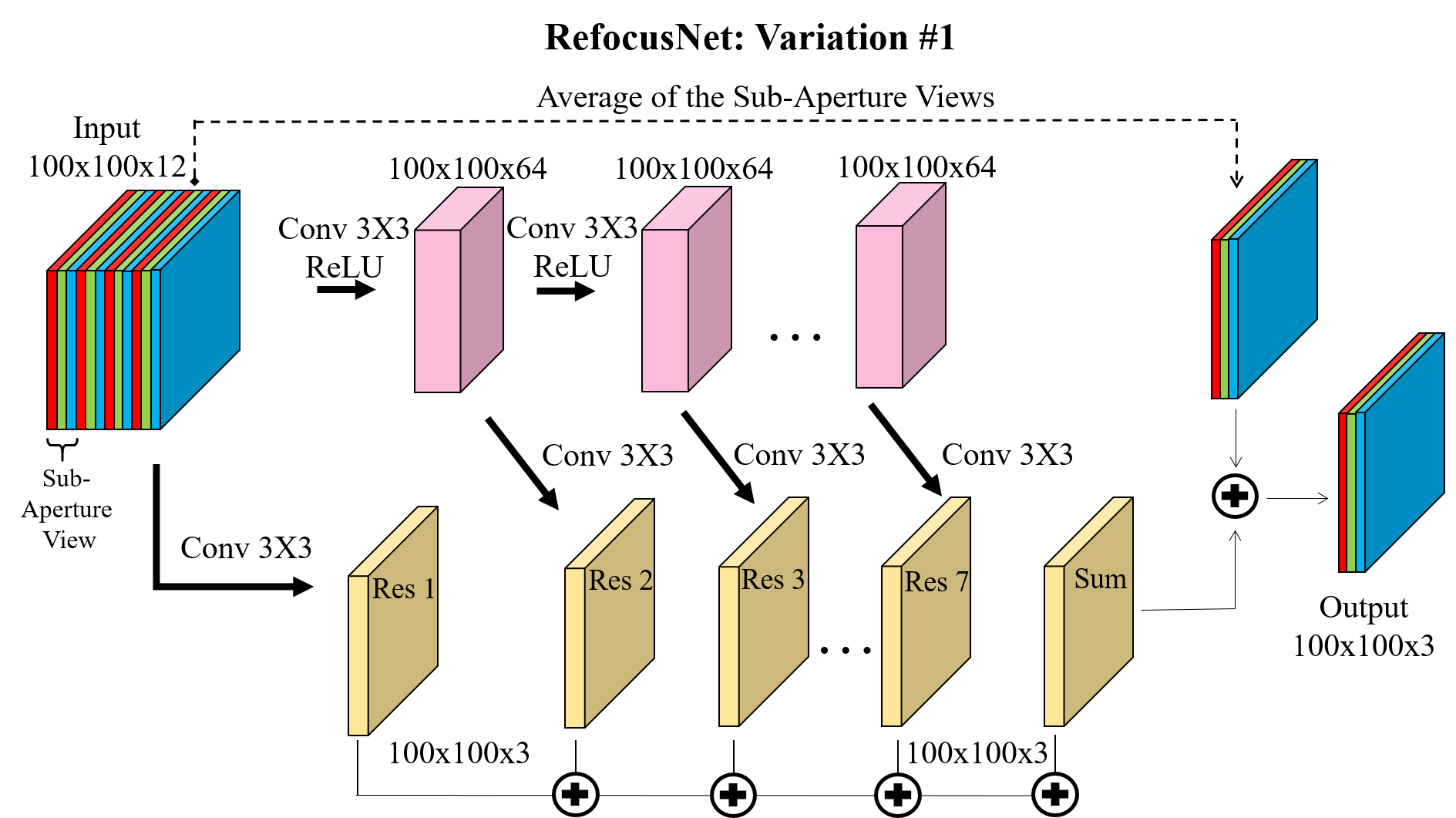}
\caption{RefocusNet variation 1. Performance [PSNR/SSIM]: clean data - 46.21/0.997, noisy data - 39.95/0.989}
\label{fig:4}
\end{figure}

As mentioned in section 4 in the main paper, part of our RefocusNet was based on an existing architecture named DenoiseNet (\cite{remez2017deep, remez2018deep}). We will now explain the motivations that led us to build the final architecture. At first, we only added a second input to the architecture: the average of the four input views, which makes the network robust to different focus sizes. A diagram of this architecture is presented in figure \ref{fig:4}. While the PSNR results on clean data were great - 46.21 dB, the results on noisy data were not satisfying - 39.95 dB. We figured that the noise appearing in the second input is the cause. 

\begin{figure}[H]
\centering
\includegraphics[width=\linewidth]{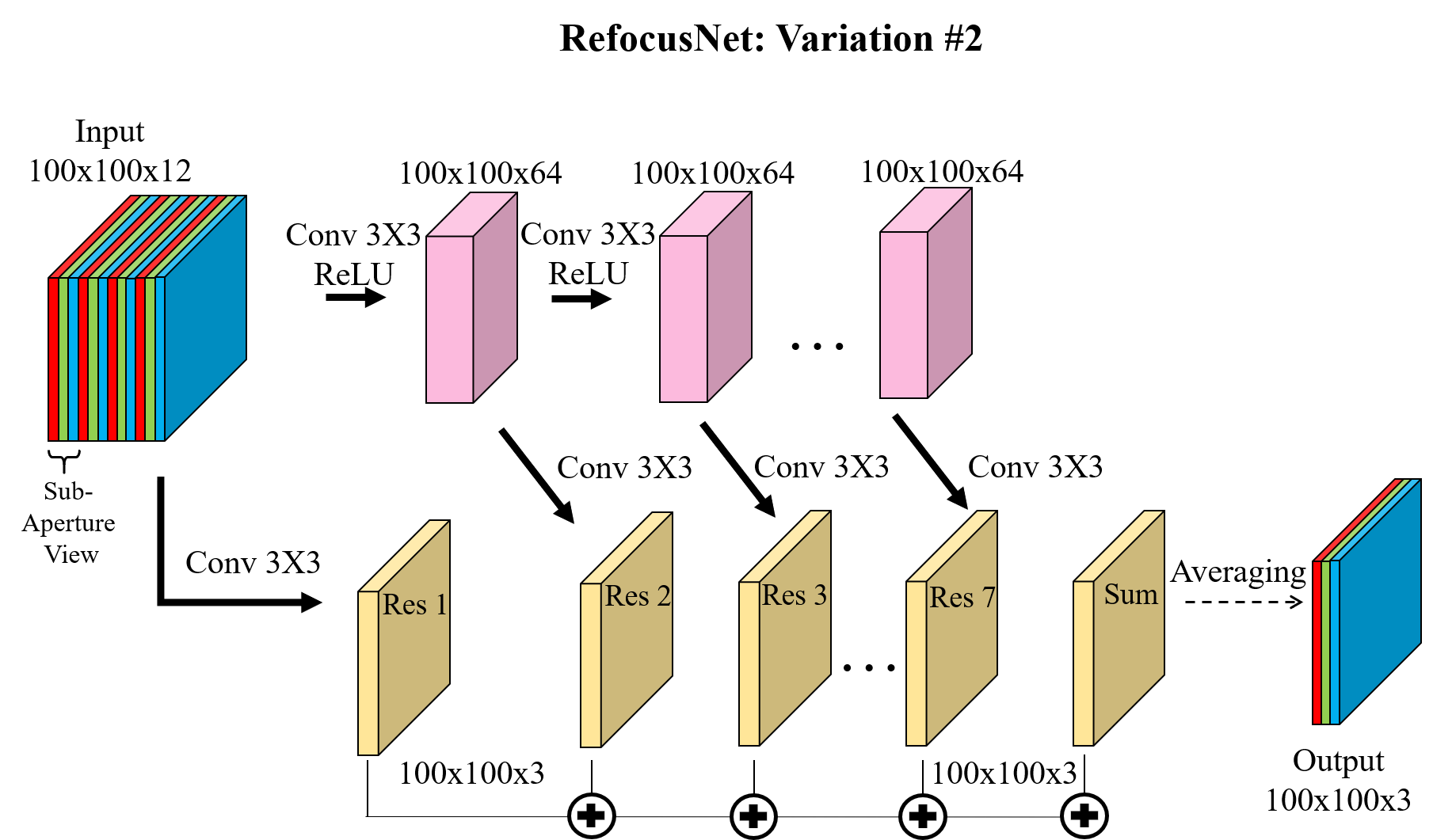}  
\caption{RefocusNet variation 2. Performance [PSNR/SSIM]: clean data - 45.85/0.997, noisy data - 40.17/0.990}
\label{fig:5}
\end{figure}

Next, we decided to take this averaging from the sum of residuals, figuring that at this point, the network probably cleaned the noise. A diagram of this architecture is presented in figure \ref{fig:5}. Although the PSNR results on noisy data improved to 40.17 dB, the results on clean data degraded to 45.85 dB. 

Finally, we decided to keep using the average calculate from the sum of residuals, but also to add an additional convolution layer on the sum of residuals and then sum these both layers together. The PSNR results on clean data improved back to 46.15 dB, and results on noisy data also improved to 40.96 dB. This is the final architecture shown in the paper.

This study shows the contribution of this specific architecture to our problem: it preserves the parts in the refocused image that we wish to preserve (the parts where the object is in focus), and repairs the parts that need smoothing e.g. the parts with the ghosting artifacts.  


\subsubsection{\textbf{Noisy Data}}
We performed an ablation study on noisy data, using Gaussian noise with increased standard deviations. The network was trained using noise augmentation, with noise standard deviation distributed uniformly at random from the range [0, 0.08]. During test time, we used standard deviation ranges from [0, 0.2]. Table \ref{table:6} shows the results of RefocusNet and the results of Kalantari \textit{et al.} \cite{kalantari2016learning}, divided into two sections: results on noisy data with standard deviations that the network was trained on ([0,0.08]) and results on standard deviations that were not seen during training ([0.1, 0.2]).

As expected, the performance degrades when increasing the noise's standard deviation, but our network still performs well on noises that were introduced to it during inference for the first time. Also, our method outperforms Kalantari \textit{et al.} \cite{kalantari2016learning} at all standard deviations except for the largest (0.2), and the difference is decreasing when increasing the standard deviations (as it becomes farther apart from the range seen during training).

It is important to note that the original algorithm of Kalantari \textit{et al.} \cite{kalantari2016learning} does not pad the convolutional layers, and also crops the results to avoid border effects created by the receptive fields during training. Thus, they crop the output images by 22 pixels from each side in both dimensions. Yet, since the network only synthesizes the angle views and does not refocus them, the refocused result images still have border effects as mentioned in Section 5 in the paper (6 pixels from each side). In order not to lose that much information from the images, we re-trained their network with padding in each convolutional layer and canceled the manual cropping at the end of the synthesizing. We only cropped 6 pixels from each side after refocusing, as we did with our networks. This operation actually improved the performance of their algorithm (PSNR improvement of $\sim$2dB), both on clean and noisy data, but not beyond RefocusNet.

\begin{table}[H]
\centering
\begin{tabular}{|ccc|}
\hline
\multicolumn{3}{|c|}{\textbf{Standard Deviations in the range [0.02,0.08]}}                                                                                 \\ \hline
\textbf{\begin{tabular}[c]{@{}c@{}}$\boldsymbol{\sigma}$\end{tabular}} & \textbf{Ours PSNR/SSIM} & \textbf{Kalantari \textit{et al.} \cite{kalantari2016learning}  PSNR/SSIM} \\ \hline
0.02                                                                    & 44.44/0.996             & 42.13/0.988                  \\ \hline
0.04                                                                    & 43.23/0.995             & 41.36/0.985                  \\ \hline
0.06                                                                    & 42.05/0.992             & 40.45/0.980                  \\ \hline
0.08                                                                    & 40.95/0.991             & 39.45/0.974                  \\ \hline
\multicolumn{3}{|c|}{\textbf{Standard Deviations outside the range [0.02,0.08]}}                                                                               \\ \hline
\textbf{\begin{tabular}[c]{@{}c@{}}$\boldsymbol{\sigma}$\end{tabular}} & \textbf{PSNR/SSIM}      & \textbf{Kalantari \textit{et al.} \cite{kalantari2016learning} PSNR/SSIM} \\ \hline
0.1                                                                     & 39.93/0.988             & 38.48/0.966                  \\ \hline
0.12                                                                    & 38.91/0.985             & 37.51/0.956                  \\ \hline
0.14                                                                    & 37.81/0.981             & 36.54/0.945                  \\ \hline
0.16                                                                    & 36.50/0.976             & 35.67/0.933                  \\ \hline
0.18                                                                    & 34.90/0.969             & 34.85/0.918                  \\ \hline
0.2                                                                     & 32.97/0.985             & 34.11/0.904                  \\ \hline
\end{tabular}
\caption{Performance of RefocusNet and Kalantari \textit{et al.} \cite{kalantari2016learning} on noisy data, divided into two sections: results on noisy data with standard deviations that the networks were trained on between the range of [0.02,0.08], and results on standard deviations that the networks were not trained on.}
\label{table:6}
\end{table}

\section{\textbf{Additional Refocusing Results}}
\begin{figure}[H] 
\centering
\includegraphics[width=0.9\linewidth]{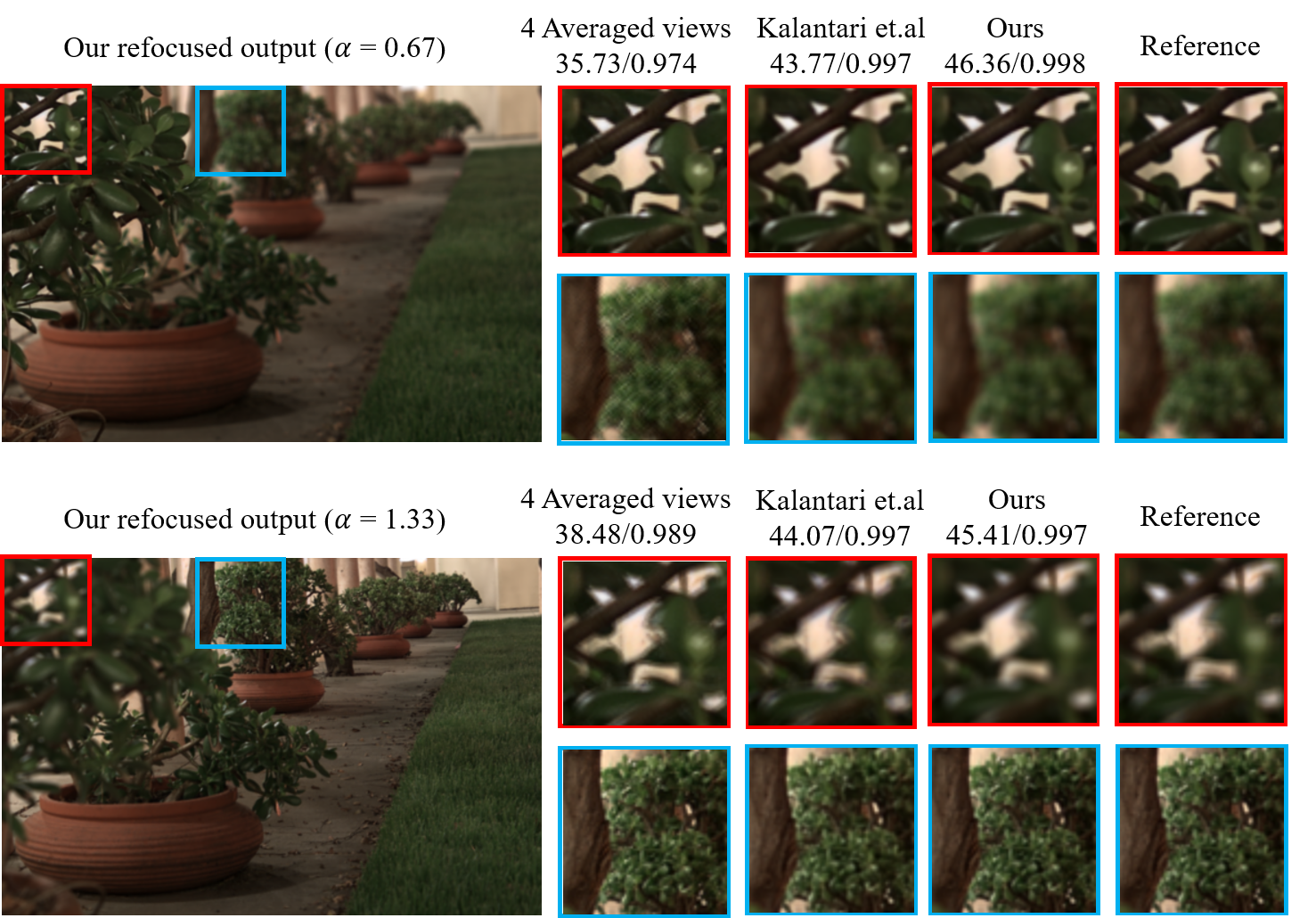}
\caption{Comparison of refocusing results  (PSNR/SSIM) between the traditional method of shifting and averaging with 4 given sub-aperture views  (\cite{ng2005light,levoy2004synthetic,vaish2004using}), Kalantari \textit{et al.} \cite{kalantari2016learning} and our RefocusNet. Our method generates a continuous refocused image, which is very similar to the reference image (calculated using 241 sub-aperture views).  However, the results of Kalantari \textit{et al.} still have some ghosting artifacts around the leaves (see the red patch of focus 1.33).  }
\label{fig:6}
\end{figure}

\begin{figure}[H] 
\centering
\includegraphics[width=0.8\linewidth]{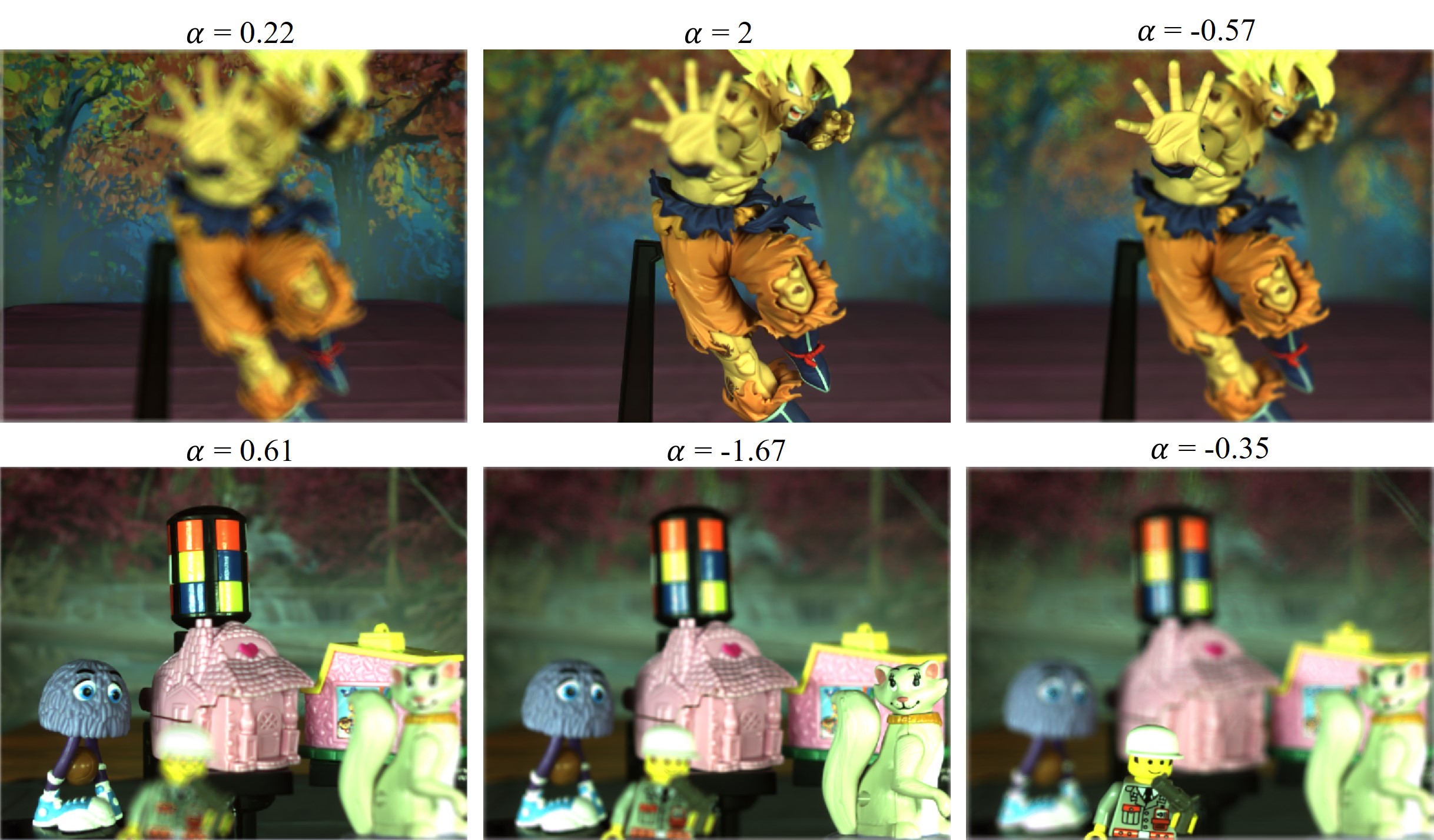}
\caption{Additional results of refocusing on images taken with our demo camera using RefocusNet. Note the exact details of the Lego-man in the bottom right image.}
\label{fig:7}
\end{figure}

\bibliographystyle{abbrv}
\bibliography{supplementary_material}